\documentclass[conference, a4paper ]{IEEEtran}
\ifCLASSINFOpdf
\else
\fi

\usepackage{algorithm}
\usepackage{algorithmic}
\usepackage{graphicx}
\usepackage{subfigure}
\usepackage{color}
\usepackage{multirow}
\usepackage{url}
\usepackage{epstopdf}
\usepackage{ctable}
\usepackage{multirow}
\usepackage{flushend}

\begin{document}
%

\newcommand*{\TitleFont}{%
      \usefont{\encodingdefault}{\rmdefault}{b}{n}%
      \fontsize{20}{20}%
      \selectfont}
     
\title{\TitleFont Top Down Approach to Detect Multiple Planes in Images \vspace{-1.5em}}
\author{\IEEEauthorblockN{Prateek Singhal\hspace{1cm}  Aditya Deshpande\hspace{1cm} N Dinesh Reddy\hspace{1cm} K Madhava Krishna}
\IEEEauthorblockA{
International Institute of Information Technology\\
Hyderabad, India}\vspace{2em}}


\maketitle

\begin{abstract}
Detecting multiple planes in images is a challenging problem, but one with many applications. Recent work such as J-Linkage and Ordered Residual Kernels have focussed on developing a domain independent approach to detect multiple structures. These multiple structure detection methods are then used for estimating multiple homographies given feature matches between two images. Features participating in the multiple homographies detected, provide us the multiple scene planes. We show that these methods provide locally optimal results and fail to merge detected planar patches to the true scene planes. These methods use only residues obtained on applying homography of one plane to another as cue for merging. In this paper, we develop additional cues such as local consistency of planes, local normals, texture etc. to perform better classification  and merging. We formulate the classification as an MRF problem and use TRWS message passing algorithm to solve non metric energy terms and complex sparse graph structure. We show results on challenging dataset common in  robotics navigation scenarios where our method shows accuracy of more than 85 percent on average while being close or same as the actual number of scene planes. 
\end{abstract}
\IEEEpeerreviewmaketitle

\section{Introduction}

\begin{figure}[!t]
\centering
\begin{tabular}{rr}
\includegraphics[width=0.22\textwidth]{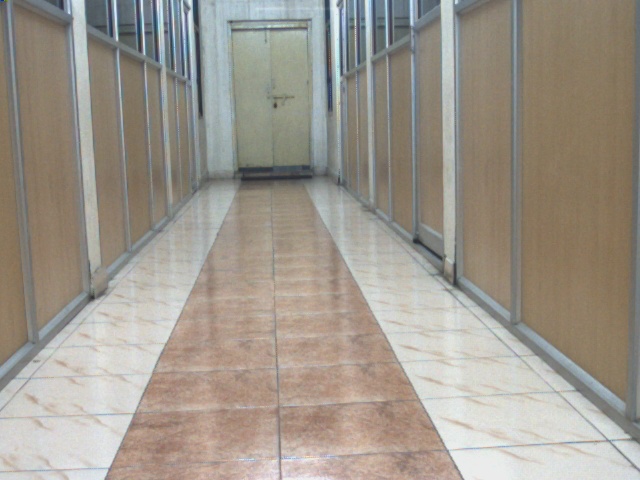} &
\includegraphics[width=0.22\textwidth]{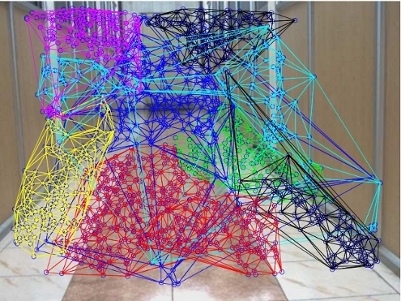} \\
\includegraphics[width=0.22\textwidth]{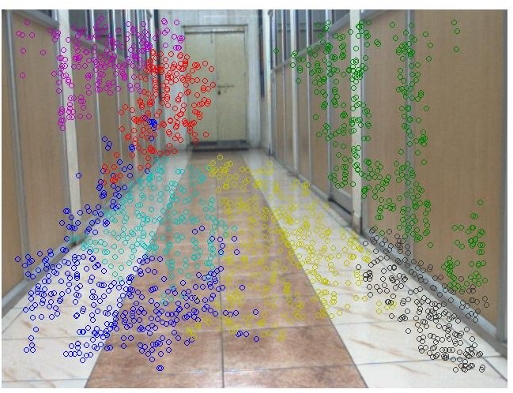} &
\includegraphics[width=0.22\textwidth]{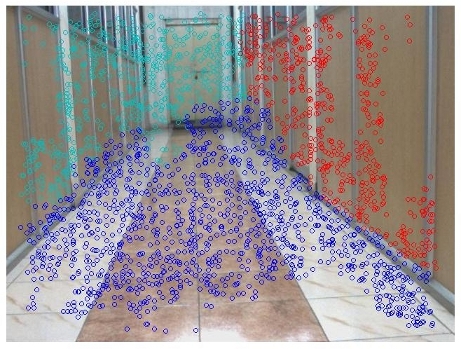} \\
\end{tabular}
\caption{ Multiple Plane detection from images.{ \bf Top Left} Initial Image to be segmented. {\bf Top Right} Initial planar patches detected from ORK. {\bf Bottom Left} Refined patches after distance based segmentation of planes {\bf Bottom Right} Multiple planes detected corresponding to scene planes after optimal MRF labelling.}
\label{figureOurResults}

\end{figure}

Detecting multiple planes in images is a challenging 
problem. If done accurately, it can provide strong cues 
to efficiently perform many vision tasks. Previously,
K\"{a}hler and Denzler~\cite{KahlerVISAPP,KahlerTPAMI},
Zhou et al.\ \cite{ZhouPlanarSfM} demonstrated the 
use of multiple planes for 3D reconstruction. Zhou et 
al.\ also exploit multiple planes for video stabilization 
\cite{ZhouJM13}. Pham et al.\ develop an augmented reality 
application using multiple planar regions of images 
\cite{PhamRCM}. Kumar and Jawahar use multiple planes to 
guide camera positioning for robot manipulators 
\cite{kumar2006robust}. To find planes, these methods 
use: $(i)$ manual annotation \cite{KahlerTPAMI} or, $(ii)$ 
iterative RANSAC methods \cite{KahlerTPAMI,ZhouPlanarSfM} 
or, $(iii)$ fit planes in the 3D reconstructed output 
(easier than detecting the planes in images) \cite{ZhouJM13}. 
We leverage recently developed multiple structure detection 
methods and build a sophisticated approach to identify 
multiple planes given two images. Our approach gives good 
results on challenging datasets (few are shown 
in Figure \ref{figureOurResults}), on which the methods 
discussed above either fail or work, but use extra 
information in the form of 3D reconstruction.

In most of the applications discussed above viz.\ 3d reconstruction, 
augmented reality, video stabilization etc.\ more than 
one image of the planar scene is available. In this 
paper, we focus on the problem of detecting multiple 
planes given two images (i.e. an image pair). Given 
this setting we can compute stable features such as 
SIFT, SURF and also find feature matches between the 
two images. These matches can then be used to compute 
a {\em homography}, which encodes the transformation 
between planar region seen in the image pair. There are 
robust RANSAC based methods, for estimating a single 
homography, a survey of these methods can be found in 
\cite{Agarwal05asurvey}. Estimating a single homography 
typically gives us only the single most dominant planar 
region in the image. To find multiple planar regions, we 
need to estimate multiple homographies from the feature 
matches.

Initially, multiple homography estimation was performed
using iterative methods. These methods eliminated the 
inliers of homography estimated for current iteration
and again performed homography estimation on remaining
matches. They typically need a priori knowledge of number 
of homographies/planes, otherwise we do not know when to 
terminate the iterative procedure. Also, the errors get 
compounded if a few wrong inliers are removed in the 
initial iterations and we end up achieving spurious 
results. More recently, sophisticated multi-structure
detection methods have been developed by Toldo and 
Fusiello \cite{Toldo}, Chin et al.\ \cite{ChinICCV09},
Pham et al.\ \cite{PhamRCM} and Jain and Govindu 
\cite{HOCJain}. These methods bootstrap by randomly 
generating many hypotheses that fit a subset of data. 
The individual data points (in our case feature matches) 
are then associated with all the hypothesis (in our case 
homographies) that they fit well. The data points are 
then clustered into multiple structures based on the 
similarity of the set of hypothesis that they match to. 
The underlying idea is data points belonging to the same 
structure will show a preference to the same hypotheses 
from our initial sampled set. These multiple structure 
detection methods are domain independent and they show 
remarkable results when applied to problems such as 
multiple 2d-line fitting, multiple 2d-circle fitting, 
multiple 3d-plane fitting etc.\ When we use these methods 
for our problem of {\em multiple homography estimation}, 
we see that these methods at best provide us with multiple 
small planar patches (after some post processing), far 
exceeding the number of planes in our image pair. We use 
realistic scenes for all our experiments. This difference 
in performance is a result of sampling homographies that 
fit nearby points and also, there being more ambiguity 
in the problem of homography estimation as compared to 
curve fitting.

After detecting multiple structures, typically some 
hueristic merging methods are used. These methods merge 
two homographies provided the residual error after 
merging is small. These methods do not work well in 
practice, this is demonstrated by the experiments done 
in Section III D. In our work, we develop a 
novel alternative to merge the planar patches 
output by multiple structure detection methods. We use 
homography decomposition to associate estimated 
homographies of planar patches to their normals in 3d world. 
At this stage, some planar patches have incorrect normals 
and cannot be trivially merged. We propose an MRF model using TRWS to 
achieve the merging. Each feature match is assigned a random 
variable, which can take labels corresponding to initial 
patches. We arrive at an optimal labelling of the feature 
matches by minimizing an energy function defined over these 
random variables. Using 
texture and locally computed normals in our energy 
minimization function, we show that we are able to assign 
correct labels to feature matches that span large scene 
planes. This is primarily because we incorporate local 
normals in our MRF formulation, majority of which are 
correctly oriented despite the normal of the entire patch 
being incorrect. Also, our smoothness term in the MRF 
formulation ensures that labels assigned to a feature match 
are consistent with its neighboring matches. As discussed in
section IV, we show good results on challenging 
datasets and compare our performance to other state-of-the-art 
multiple plane detection methods.

\section{Related Work}
\label{sectionRelatedWork}
\begin{figure}[!t]
\centering
\begin{tabular}{cc}

\includegraphics[width=0.22\textwidth]{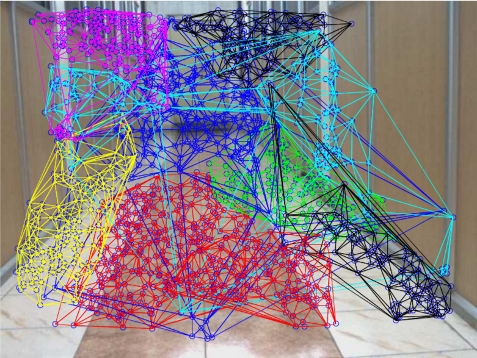} &
\includegraphics[width=0.22\textwidth]{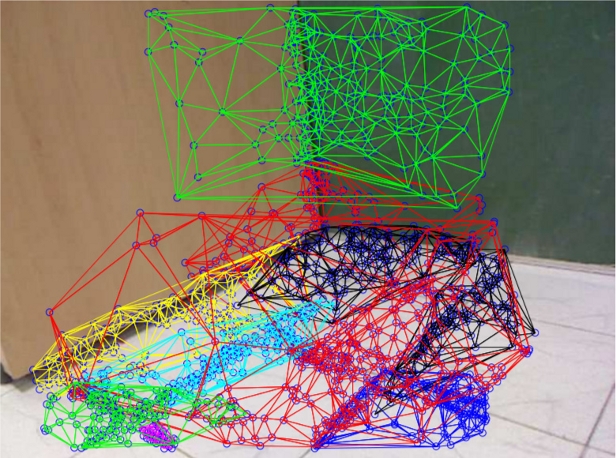} \\
\end{tabular}
\caption{Scene planes detected for {\it corridor, corner} dataset
respectively (from left). Connected meshes with the same color 
indicate matches that are grouped together as belonging to one
plane by \cite{ChinICCV09}.}
\label{figureORKResults}

\end{figure}

RANSAC based homography estimation methods have been
extended to the problem of detecting multiple planes
by removing inliers and re-estimating new homographies
iteratively. Further, Zuliani et al.~\cite{multiRANSAC} 
developed a multi-RANSAC algorithm that is capable of 
estimating all homographies simultaneously. These methods 
do not work well in practice and also, need additional 
knowledge of number of planes.

\subsection{Methods using J-Linkage}

Recently, sophisticated algorithms which do not require 
prior knowledge of number of planes have also been 
developed. Toldo and Fusiello \cite{Toldo} develop one 
such algorithm called {\em J-Linkage}. For homography 
estimation, J-Linkage starts by generating $M$ homographies 
from randomly sampled nearby feature matches. For each 
feature match, a preference set of the homographies (out 
of the $M$) that fit to the match within a threshold 
$\epsilon$ is created. A clustering step that starts 
with all matches in separate clusters is performed. The 
clustering step iteratively merges the feature matches 
that have similar preference set using the Jaccard 
Distance measure ($d_{J}(X,Y) = \frac{|X \cup Y| 
- |X \cap Y|}{|X \cup Y}$). This clustering step proceeds 
till the minimum $d_{J}$ is $1$, i.e. the preference 
set of all clusters have no more overlap. The outlier 
feature matches are also present as small clusters and 
these are eliminated by setting a threshold on cluster 
size. 

Fouhey et al.~\cite{Fouhey} observe that J-linkage 
uses only nearby feature matches to generate initial 
homographies. This is done to ensure that the computed
homographies correspond to real scene planes. Though,
the downside of such an approach is that the 
homographies output by J-linkage are also locally 
optimal and do not fit a large scene plane. The
scene plane is output as multiple small planar 
patches by J-linkage. Fouhey et al.~\cite{Fouhey}
solve this by continuing to cluster the matches 
(after J-linkage) using the distance measure:

\begin{center}
$d_{F}(X,Y) = \frac{1}{|X \cup Y|}\sum \limits_{c \epsilon X,Y} err_{H'}(c)$
\end{center}

\subsection{Methods using Ordered Residual Kernel}
\label{sectionORK}
Another category of multiple structure detection algorithms
based on ordered residues have been formulated by Chin et al.
\cite{ChinICCV09}. For the problem of multiple homography 
estimation, again these methods start by randomly sampling
$M$ homographies. Residues ($x' - Hx$) are then computed for 
each feature match and homographies are ordered on the
basis of increasing residue for each feature match. Thus, given 
a data point $\theta_{i}$ we obtain ordered homographies 
$\lambda^{i}_{1}$ (minimum residue) to $\lambda^{i}_{M}$ 
(maximum residue):

\begin{center}
$\tilde{\theta_{i}} = \{ \lambda^{i}_{1}, \lambda^{i}_{2}, \lambda^{i}_{3}, ...,  \lambda^{i}_{M-1}, \lambda^{i}_{M}\}$
\end{center}

An Ordered Residual Kernel (ORK), $k_{r}$, is defined between two
data points:

\begin{center}
$k_{r}(\theta_{i1}, \theta_{i2}) = \frac{1}{Z} \sum \limits_{t=1}^{M/h} z_{t} k^{t}_{\cap}(\tilde{\theta_{i1}}, \tilde{\theta_{i2}})$
\end{center}

$k^{t}_{\cap}$ is the Difference of Intersection Kernel 
(DOIK) defined on the homographies ordered by residues 
($\tilde{\theta_{i1}}$ and $\tilde{\theta_{i2}}$) of
the two data points.

\begin{center}
$k^{t}_{\cap} = \frac{1}{h} (|\tilde{\theta_{i1}}^{1:\alpha_{t}} \cap \tilde{\theta_{i2}}^{1:\alpha_{t}}| 
		- |\tilde{\theta_{i1}}^{1:\alpha_{t-1}} \cap \tilde{\theta_{i2}}^{1:\alpha_{t-1}}|)$
\end{center}
\begin{figure}[!t]
\centering
\begin{tabular}{cc}
\includegraphics[width=0.22\textwidth]{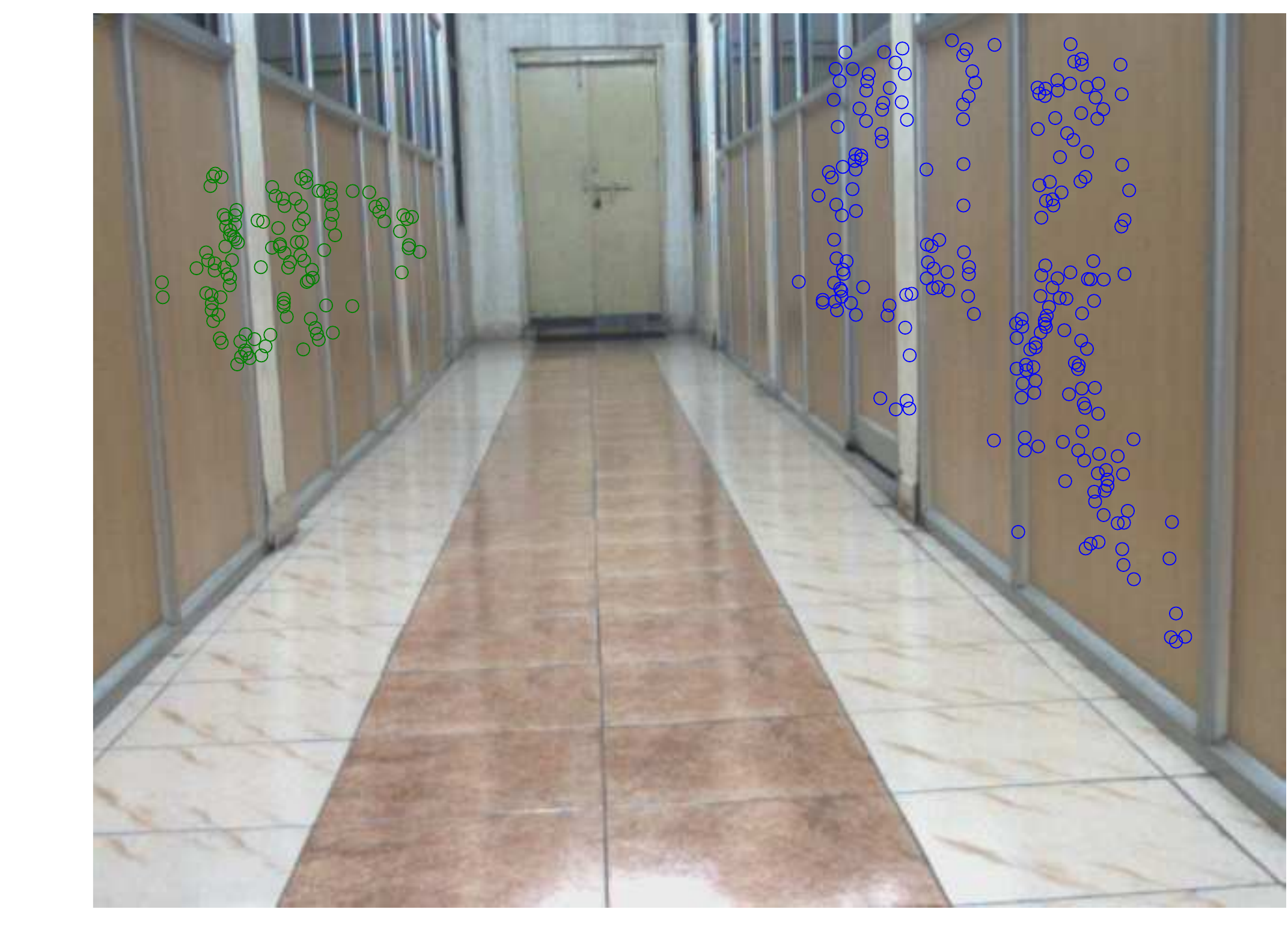} &
\includegraphics[width=0.22\textwidth]{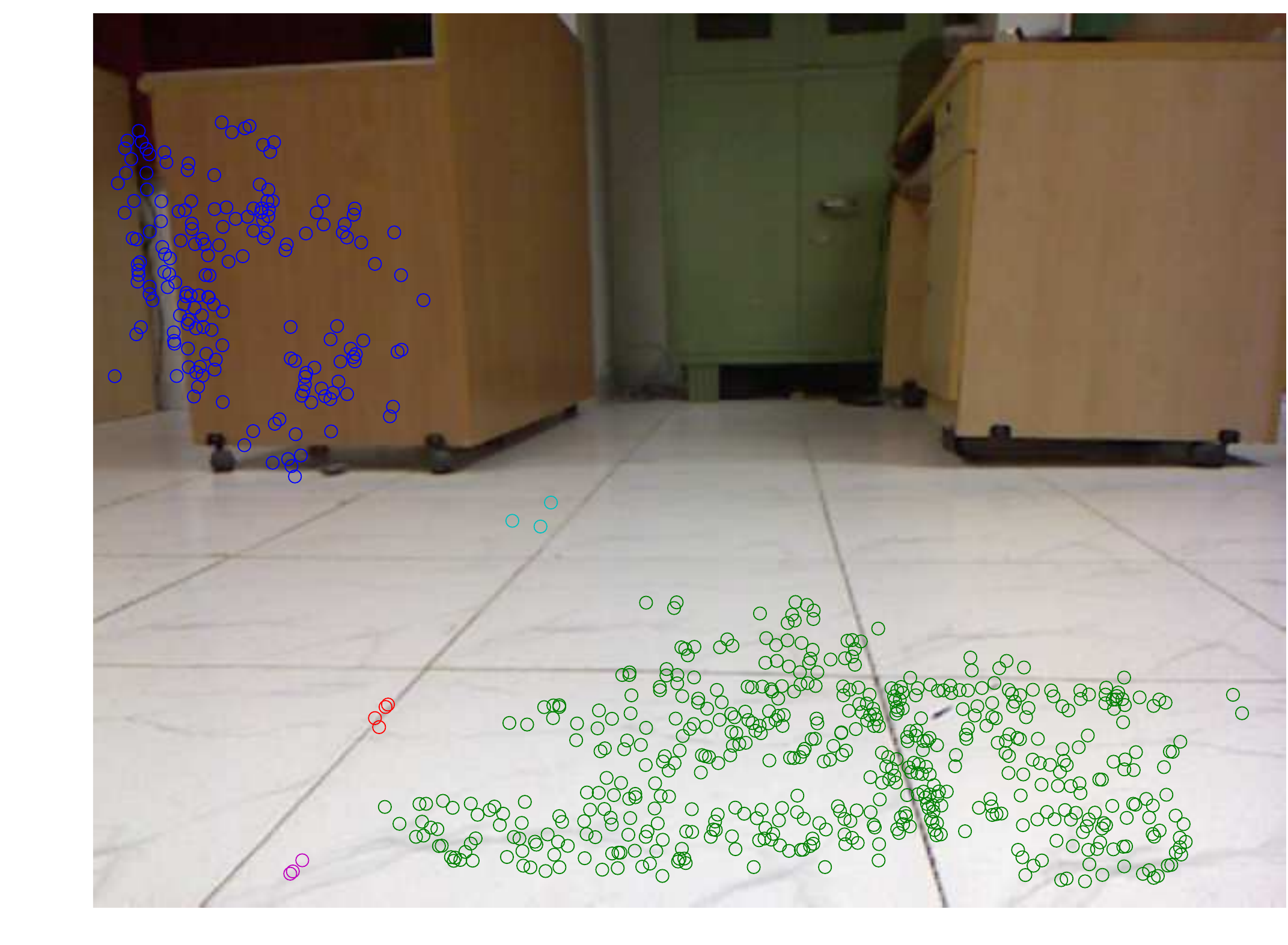} \\
\end{tabular}
\caption{Planar patches obtained after the distance
based refinement step for corridor and corner dataset. In comparison to Fig 2, the 
images show that our distance based refinement step is able to cut the 
detected scene planes (that spanned two or more real world planes) 
into smaller planar patches (that span only a single scene plane). }
\label{figureDelaunay}
\end{figure}

The ORK is a weighted sum of the difference in number 
of intersecting homographies taken over some step size 
$h$. This kernel is a valid mercer kernel and induces 
a mapping of the data points to a Reproducing Kernel 
Hilbert Space (RKHS). Chin et al.\ show that data points 
belonging to the same structure form clusters in this 
RKHS. They use kernel PCA and spectral clustering to 
detect these structures. To minimize the number of 
structures detected after this step, they also give a 
structure merging scheme. This merging scheme 
sequentially merges structures if the overall residue 
after merging is below a threshold. The merging 
continues till all the data can be explained 
satisfactorily by identified structures (i.e. sum of 
all residues is bounded). 

Like Fouhey et al.\ \cite{Fouhey}, we propose an 
alternative scheme for merging structures output
by the multiple structure detection methods.We compare the performance of our 
approach to Fouhey et al.\ \cite{Fouhey} and the 
domain independent merging scheme given by Chin et 
al. \cite{ChinICCV09}.

\section{Our Approach}

The following sections describe the different steps
of our approach.

\subsection{Initial Planar Patch Estimates}

Given two images we compute SURF features and obtain
feature matches. On obtaining these matches, we 
use the multiple structure detection method described
in section \ref{sectionORK} for estimating multiple
homographies and hence, multiple scene planes. We perform
the steps of: $(i)$ computing ordered residues of feature 
matches, $(ii)$ using the ordered residual kernel, $(iii)$ 
performing kernel PCA and spectral clustering. 
We avoid using the cost of residues based structure merging 
scheme given by Chin et al. \cite{ChinICCV09}. We 
justify in Section III-D that such schemes do not work well 
in practice. Refer \cite{ChinICCV09} for more details of 
this method. Figure \ref{figureORKResults} shows the scene 
planes that are found by using this method on two datasets: 
$(i)$ {\it corridor}, consisting of $4$ planes and $(ii)$
{\it corner}, consisting $3$ planes representative 
of a corner of a cuboid. As seen in Figure 
\ref{figureORKResults}, the detected planes have
two major problems:

Firstly not all detected planes correspond 
to a planar patch. For example, the blue mesh in {\it 
corridor} dataset shows that the detected plane spans left 
as well as right plane, the $2$ red meshes in {\it corner} 
dataset show that $2$ detected planes span ground and left, 
right and left planes respectively.Also,the number of planes detected by the 
multiple structure detection method exceeds the number 
of true scene planes.

\noindent
 \begin{figure}[!t]
\centering
\begin{tabular}{rr}
\includegraphics[width=0.22\textwidth]{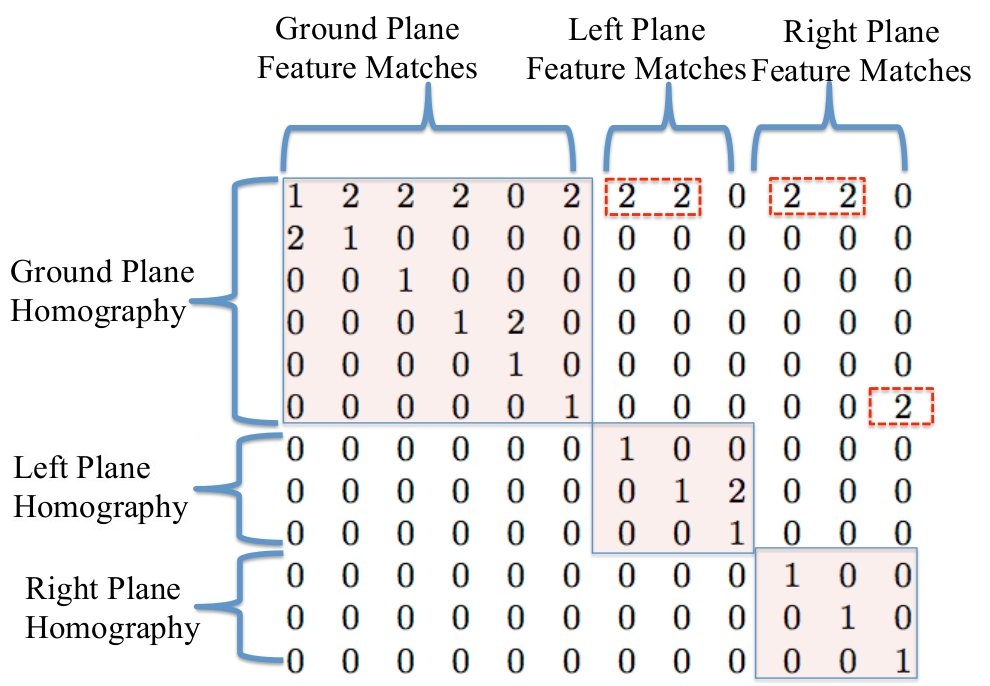} &
\includegraphics[width=0.22\textwidth]{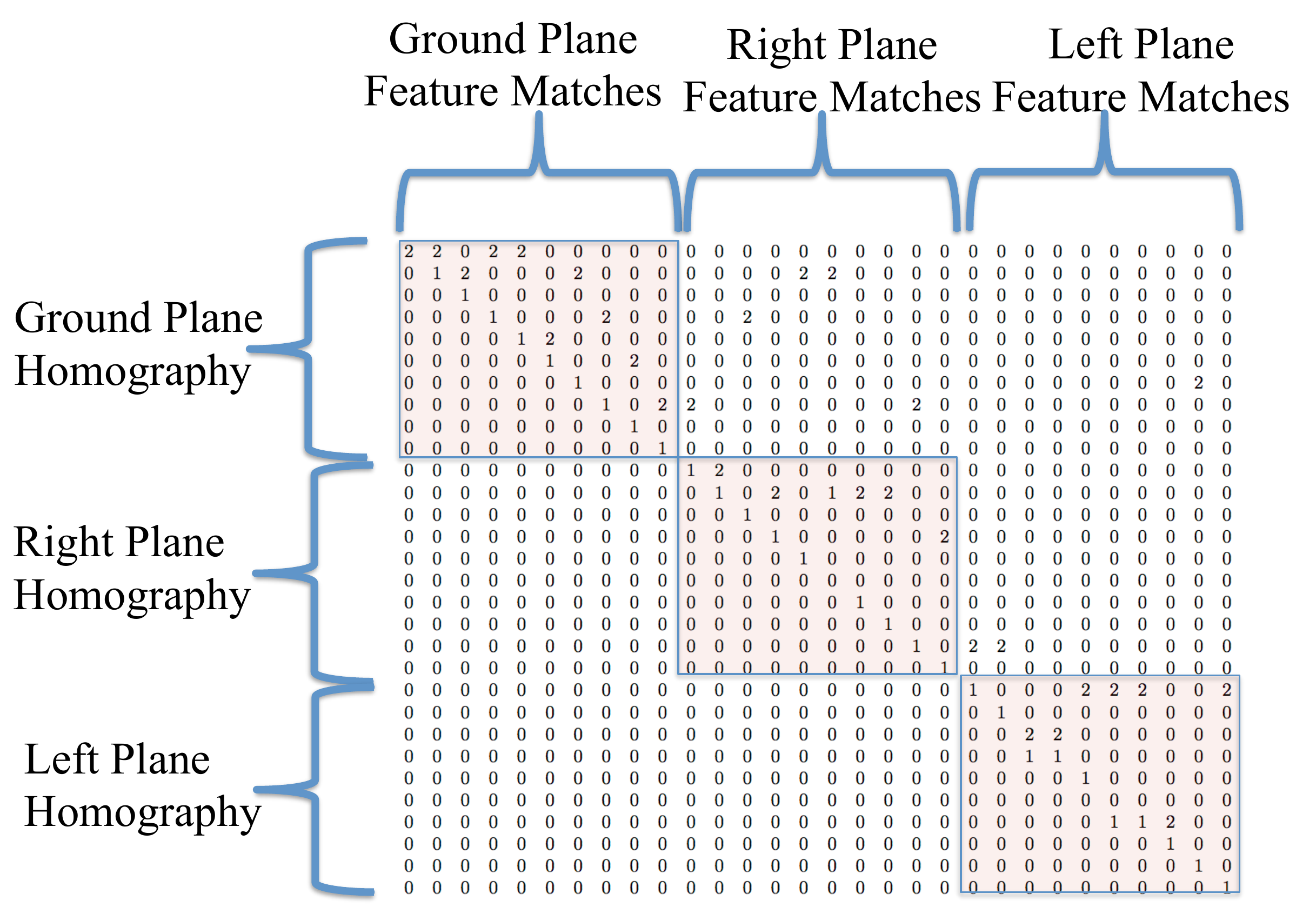} \\
\end{tabular}
\caption{ The minimum and second minimum after 
applying the homography of one plane to feature matches of
the other.}
\label{figureHomogExpt}
\end{figure}
We solve these problems using the following pipeline:

\begin{itemize}
\item Distance based refinement of detected planes. 
\item Computation of plane normals.
\item MRF based optimal labelling due to local and texture constraints.
\end{itemize}	 

\subsection{Distance Based Refinement of Detected Planes}
In this step, we first perform a delaunay triangulation on the
feature matches. If the detected plane has feature matches
from multiple scene planes, chances are these would be at
a larger distance from each other. We set a threshold on 
distance and cut the delaunay mesh into smaller meshes (when 
any side exceeds the distance threshold). This step, to some 
extent, separates the planes output by multiple structure 
detection method into smaller planar patches. These smaller
planar patches have stable properties as compared to detected
planes which spanned multiple true scene planes. The results of 
performing this step on {\it corridor, lab} dataset are 
shown in Figure \ref{figureDelaunay}.

\subsection{Plane Normal Computation}
Planar points between two images are related by homography. 
Homography is a relation between the plane 
and the relative pose between the images which can be 
decomposed \cite{ZhouPlanarSfM} to find the plane 
normals.  

\hspace{3cm}$H= (R + \frac{TN^t}{D})$ \\

where N is the plane normal and D is the perpendicular distance between the plane and camera. We  have found that in cases of perspective motion between the  camera and the plane, in presence of multiple planes, the decomposition is mostly erroneous due to the bilinear nature of the normal and translation term .We also discard planes formed by less than 10 points as they are mostly erroneous.

\subsection{Residue Based Merging of Detected Planes}
Typically methods of detecting multiple homographies,
including Chin et al.\ resort to merging the detected
planes by applying the homography of one to other. The
merging is done, if the residue after applying the
homography of another plane is below a threshold. We
design an experiment, where we perform the above on:
$(i)$ planar patches out put by Chin et al.\ before
doing the merging and $(ii)$ on planar patches that are 
manually marked, these planes have desirable properties
(viz. they span entire planar region). As shown in Figure
\ref{figureHomogExpt}, for $(i)$ we have $6$ ground, $3$ left
and $3$ right planes. For $(ii)$, we have use $10$ ground,
$10$ left and $10$ right planes. We create a matrix where
the rows indicate the homography taken and columns indicate
the plane to which it is applied. We mark out the first
and second minimum residues by $1$ and $2$ in this matrix.
The second minimum residue will dictate the merging in
residues based merging approach. In ideal conditions, all 
the $1$'s and $2$'s should lie in the shaded regions of 
Figure \ref{figureHomogExpt}. This is the case for 
controlled experiment $(ii)$, but not for $(i)$ 
(5 out of 12 planes marked in red have second minimum 
 residues for incorrect homographies). Since our 
experiments show that using residues alone is not 
sufficient, especially for multiple structure detection
methods like Chin et al., we develop other cues that can 
be exploited to achieve merging.

\subsection{Top Down Approach} 
Based on the discussion above, we propose 
exploiting the local consistency constraint (viz. same
direction of normals, similar texture etc.), that should 
hold for a planar patch, to merge scene planes. We 
consider each feature match in its local neighbourhood 
($k$ nearest neighbours in detected planar patch) to form 
a small local planar patch. This is in cognizance to local 
patches embodying a scene plane \cite{Fouhey}. With our 
decompostion to local planar patches, we can now compute 
local surface normal for each such planar patch. The 
consistency in orientation of local normals spanning a 
single scene plane is an important cue, so is its texture. 
We use these cues to refine and recompute association of 
feature matches to the detected planes. Note that we do an
MRF optimization on a sparse graph of only feature 
matches and not a dense graph of all pixels. Such dense 
graphs are common in image segmentation literature. We 
call our approach a {\it top-down approach}, because we 
resort to feature matches and local planar patches to merge
detected planes, after having performed one step of planar
patch detection (using multi-structure detection 
methods). For our approach it is necessary to perform
an initial multiple structure detection step. Because
we assume that the true structure is present in the
output of multiple structure detection method and develop
a method to merge the detected structures to these
true structures.

\subsection{Graph Optimization}
The feature matches are connected to
form a graph using delaunay triangulation. This graph 
structure is formulated as Markov Random Field where 
the goal is to assign each feature match a maximal 
posterior probability (MAP) label. This label is one
out of the detected planes by multiple structure
detection step. As a common practice [16], instead of direct probability maximization, we minimize the energy as discrete labelling problem on the graph in the form of Eqn1.

\begin{eqnarray}
•E_{MRF} = \sum_X E(p, {l \epsilon L}) + \sum_x \sum_{q\epsilon N(X) } E(p,q)
\end{eqnarray}
In the above equation, L = $(p^1..p^n)$ is the set of 
labels where $n$ is the number of initial planar patches obtained 
from initial planar patch estimate step. The set $N(X)$ is the 
neigbourhood of the node X. The $E(p,l)$ defines the unary 
energy potential. It determines the likelihood of the feature
match corresponding to a scene plane labelled $l$ output by
multiple structure detection step. $E(p,q)$ defines the pairwise 
energy potential which represents the graph similarity of the 
neighbourhood.

\subsection{Unary Energy Term}
Unary energy is defined by us as a sum of energies 
relating to {\it residues of local planar patch} and {\it normal 
similarity between feature and plane}.
\begin{equation}
E_{unary}= E(X)_{normal} +  E(X)_{residual}
\end{equation}
 \begin{figure*}[!t]
\centering
\begin{tabular}{cccc}
\includegraphics[width=0.22\textwidth]{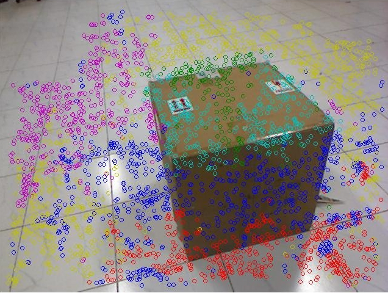} &
\includegraphics[width=0.22\textwidth]{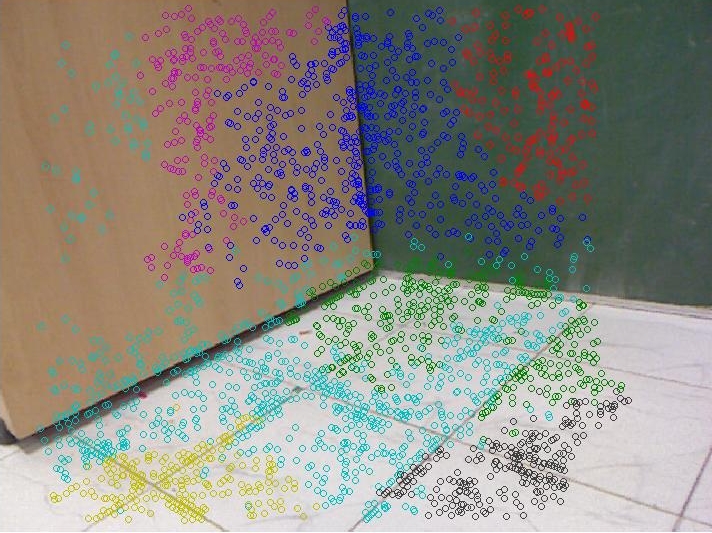} &
\includegraphics[width=0.22\textwidth]{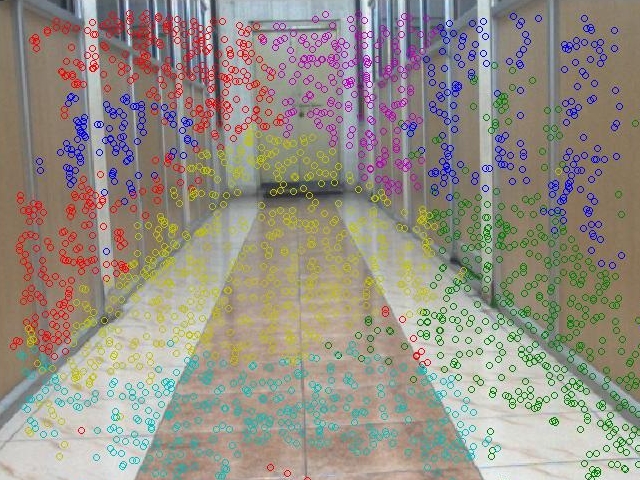} &
\includegraphics[width=0.22\textwidth]{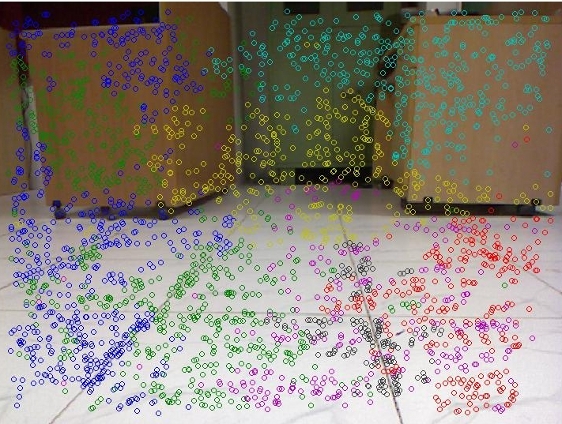} \\
\includegraphics[width=0.22\textwidth]{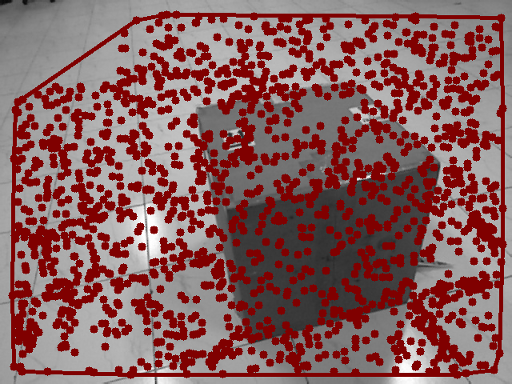} &
\includegraphics[width=0.22\textwidth]{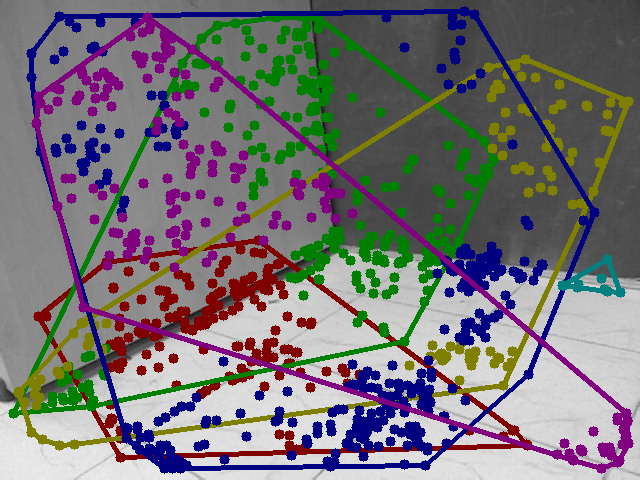} &
\includegraphics[width=0.22\textwidth]{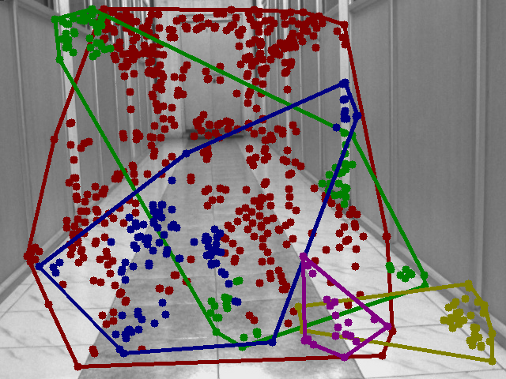} &
\includegraphics[width=0.22\textwidth]{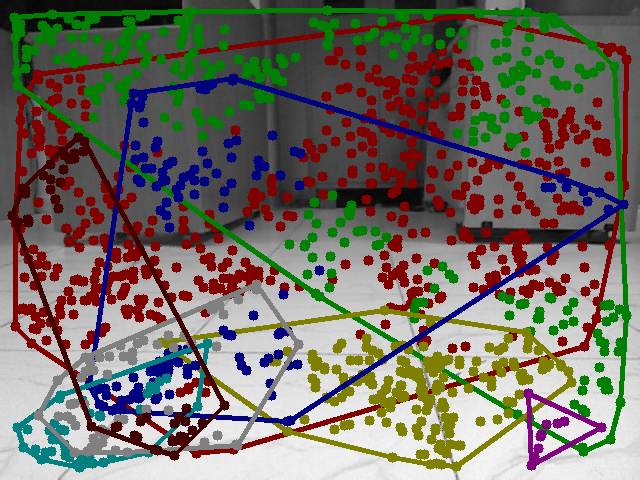} \\
\includegraphics[width=0.22\textwidth]{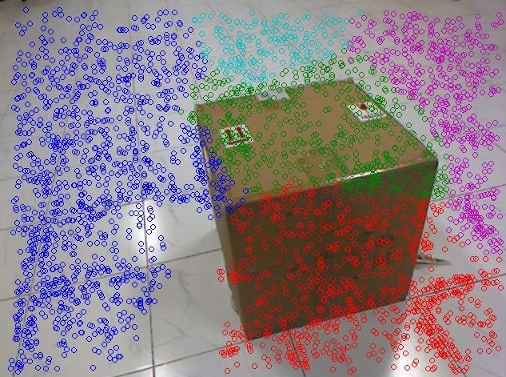} &
\includegraphics[width=0.22\textwidth]{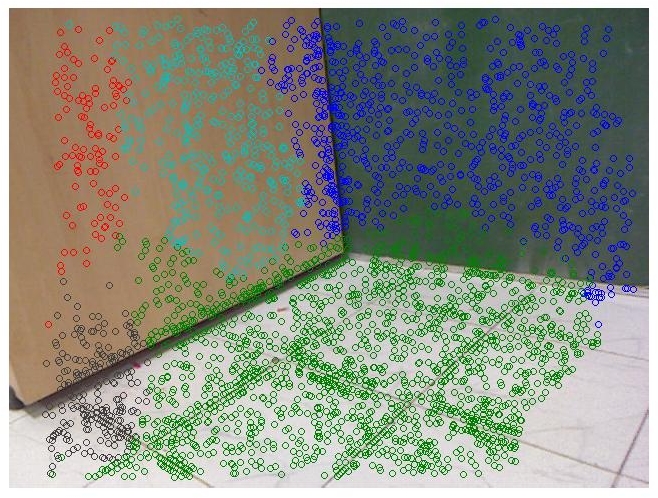} &
\includegraphics[width=0.22\textwidth]{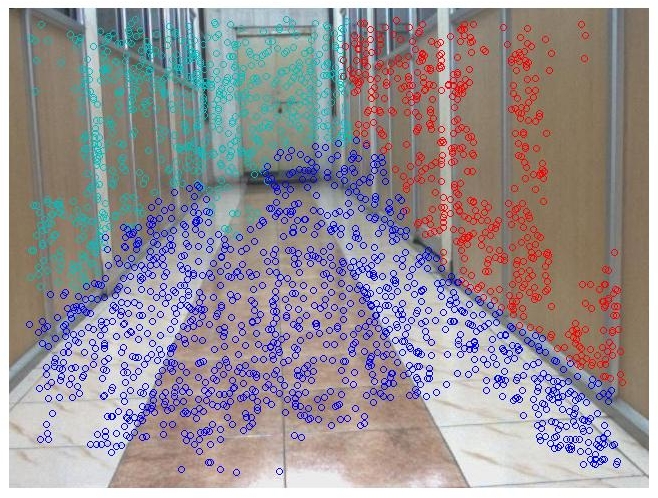} &
\includegraphics[width=0.22\textwidth]{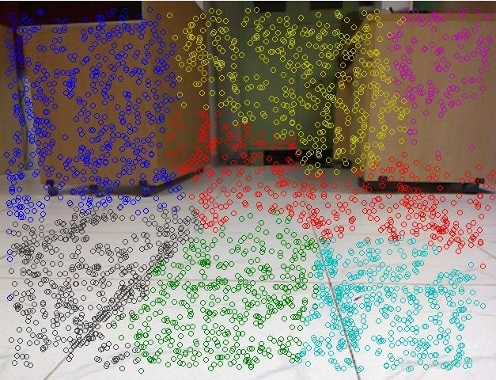} \\ 
\end{tabular}
\caption{
This figure compares the results of different multiple
plane detection methods on different datasets. Datasets - {\em
box, corner, corridor and {\em tables} (from left to right). From top
to bottom, each row corresponds to the following methods - 
Chin et al.~\cite{ChinICCV09}, Fouhey et al.\cite{Fouhey} and
our approach.}In our approach to corridor we undersegment as the normal on the perspective plane doesnt turn out well also the tracks are not good}
\label{figureFinalResults}
\end{figure*}

\subsubsection{Residue of local planar patch}
Each feature match has a local planar patch defined around it 
using $k$ nearest neighbours. Our energy is the sum of 
residues of this local patch with respect to the parameters
of the patch labelled $l$. 
\begin{eqnarray}
E(X, l)_{residual}= \sum_{X \epsilon P} ||(X'-H_{l}X)||_{L2}
\end{eqnarray}
 Here $P$ represents the local planar patch. We use $k=10$ as smaller 
patches can be erroneous while decomposition.These nearest neighbours 
are members of the same planar patch found initially as this implies 
that they are part of a larger plane rather than some local surface fit 
to a scene.
\subsubsection{Normal Similarity measure}
 The local normal of each feature match should ideally be 
aligned with the normal of the scene plane. So the energy term for each 
feature decreases if it aligns with the plane labelled $l$.
\begin{eqnarray}
E(X,l)_{normal}= (1-\frac{N_{X}.N_{l}}{|N_{X}||N_{l}|})^2
\end{eqnarray}
\subsection{Pairwise Energy Terms}
 Pairwise energy defined by us consists of three 
terms {\it similarity measure, mutual plane consistency} and 
{\it texture similarity}.
\begin{equation}
E_{binary}= {\lambda_1}E(X,Y)_{sm} + {\lambda_2}E(X,Y)_{mp} + {\lambda_3}E(X,Y)_{ts} 
\end{equation}    

where $\lambda_1$ ... $\lambda_n$ are the weights of pairwise terms. 

\subsubsection{Similarity Measure}
 We use the standard Potts Model where the neighbouring 
edges with different labels are penalized. Initially each feature 
is assigned the label of the initial planar patch to it belongs.

$$
E(X,Y)_{sm} = \left\{ \begin{array}{rl}
 1 &\mbox{ if $p_X\neq p_Y $} \\
 0 &\mbox{ otherwise}

       \end{array} \right.
$$
\begin{equation}
\end{equation}

\subsubsection{Mutual Plane Consistency}
Neighbouring features should have similar surface normals, utilizing 
this constraint we find the measure. 
\begin{equation}
E(X,Y)_{mp}= (1-\frac{N_{X_p}.N_{X_q}}{|N_{X_p}||N_{X_q}|})^2
\end{equation}
\subsubsection{Texture Similarity}
This measure takes into account the local texture between neighbours 
should be similar. Here we compare the mean of a image patch around 
each feature match with its neighbour.  We 
use a $5\times5$ patch centered at the feature. This term brings in the 
smoothness of texture across a plane, typically common in images. 
\begin{equation}
\hspace{2 cm } \mu = \frac{\Sigma_{WS} (R,G,B)}{WS}  \hspace{0.5 cm } : WS= Window Size 
\end{equation}

\begin{equation}
E(X,Y)_{cs}= ||(\mu_{p(x)} - \mu_{p(y)})||_{L2}
\end{equation}
 This combination of energy terms segments out the planes 
robustly. We choose Tree Weighted Sequential(TRWS)[15] 
message passing algorithm to  solve the optimization 
problem.   

This method is similar to Loopy Belief Propagation and solves on the 
priniciples of linear programming and its duality for NP hard MRFs. The 
method has experimentally shown better results than LBP while working 
well in cases of non metric pairwise terms. It also finds the lower 
bound of the energy which acts as a guidance for 
convergence.Thus this method provides flexibility to include varied 
and sophisticated energy terms with complex graph structures. 
\cite{Szeliski06acomparative}

\section{Experimental Results}

We evaluate our approach on various images taken in an
indoor environment. The images of our datasets and the
results of different methods for multiple plane detection
on them have been shown in Figure \ref{figureFinalResults}. Our 
images are representative of scenes that will be 
encountered by a robot in SLAM setting and also in other 
3d reconstruction methods. These datasets are challenging 
because some of our planes have multiple textures (ground
plane of corridor) and a specular reflection (glass in the 
left and right planes of corridor, ground plane of box). 
The images have been taken from a dataset captured for 
VSLAM by a Flea2 camera mounted on P3Dx robot. There is 
considerable movement between the images ($\approx$ 
around $20cm$ for corridor dataset and for others it is in 
the range of $5$ to $10$cm, with 2 to 5 degrees rotation). We 
compute SURF features and use the KLT tracker to find 
feature matches. The average number of feature matches 
found is around 3000. These features generally encompasses 
the image and the corresponding planes well. 

We compare our approach to the two approaches -- Fouhey et al.
\cite{Fouhey} and Chin et al.~\cite{ChinICCV09} -- discussed 
in section \ref{sectionRelatedWork}. For Fouhey et al. and
Chin et al.\, we use their publicly available code.For fairness we run the codes several times and the best results were taken . 
 
Using$1500$ SURF features tracked by KLT, code by Fouhey et al.\ 
took in the range of $5$ to $10$ mins per experiment. As can be 
seen from Figure \ref{figureFinalResults}, for corridor it 
finds erroneous planes. For box dataset, where there are 
parallel planes (with different textures) it labels both 
of them as same. Also, it fails to detect the front facing
plane.It performs badly in such image as the data has high amount of outliers and purely residue based merging does not help. For the corridor and lab dataset it finds multiple planes spanning other planes as well as the number of planes is grossly oversegmented . 

Taking this performance into consideration we do a quantitative and qualitative analysis of results only between ORK and our approach. We run both approaches with Multi Guided Sampling Pham et al .~\cite{PhamRCM}. Table I shows Classification error of the data and number of planes detected in the image. Our approach shows competitive results for classification error while having a lower error for most of the data. The number of planes detected by ORK is more erroneous than ours as it over segments the same plane.

\subsection{Quantitave Analysis}
 We show statistics for the {\it PS (Amount of plane detected(PD) present in scene plane(SP))} and{ \it AD (The amount of planes detected(PD) spanning the scene plane(SP))}. In an ideal situation both should correspond to 1. We show our results only for corridor and corner dataset(space constraints).Table II shows PS and AD for corridor dataset using our approach. For example first detected plane(PD I) from PS is detected primarily in second scene plane(SP II)(0.89) and from AD we can  find that PD I covers entirely SP II (0.95) thus making both the ratios close to 1. In comparison ORK (Table III )has its PD I primarily (0.986) in SP I but the PD I covers only (0.16) of SP I showing the high error ratio while classification accuracy being high. Similar analysis of the whole data shows that our method performs far better than ORK in finding true scene planes.
\vspace{-1em}
\subsection{Qualitative Analysis}
  Chin et al.\ perform better than Fouhey et al.\ for our datasets,
but there is still scope for improvement. For the box dataset, 
the topmost plane (detected in blue) has a few other planes detected
in between. Similar is the case with ground plane in the corner 
dataset, it is split into multiple planar patches.
There are also erroneous planes viz. the plane marked by
green feature matches in the corridor dataset. This plane has
matches from the left as well as ground plane. Similarly, plane
marked by blue in corner dataset spans the left and ground plane.
We are able to solve these problems in our approach, since we impose
strong locality constraints and also look at the local normals,
texture to perform merging. 
\vspace{-0.5em}
\section{Conclusion}
In this work, we develop an MRF based top down approach
to merge multiple small planar patches detected by multiple
structure detection methods. We show significant 
improvement over previous methods. This improvement results
from the fact that we bring in domain knowledge to the problem 
of multiple plane detection. Our domain knowledge is in the 
form of cues such as local normals, texture and local consistency 
of planes. We formulate an energy function using this domain
knowledge and minimize it through an MRF optimization.

\begin{table}
\caption{
error matrix
}
\centering
 \small
 \begin{tabular}{ |>{\centering\arraybackslash}m{0.4in} | >{\centering\arraybackslash}m{0.4in} | >{\centering\arraybackslash}m{0.4in} |>{\centering\arraybackslash}m{0.4in} |>{\centering\arraybackslash}m{0.4in} |>{\centering\arraybackslash}m{0.4in} |>{\centering\arraybackslash}m{0.4in} |}
 \hline
 \multicolumn{2}{ |c| }{METHOD} &  \multicolumn{2}{ c| }{OUR} & \multicolumn{2}{ c| }{ORK}\\ \cline{1-5}
\hline
 Dataset & No of Ground truth SPs & error(in '\%') & No of SPs detected & error(in '\%') & No of SPs detected\\
\hline
 Lab & 4 & {\color{blue}12.32} & {\color{red}7} & {\color{blue}18.75}  & {\color{red}7}  \\
\hline
 Corner & 3 & {\color{blue}18.9} & {\color{red}6} & {\color{blue}12.33}  & {\color{red}8} \\
\hline
 Box & 3 & {\color{blue}13.24} & {\color{red}5} & {\color{blue}13.53}  & {\color{red}6} \\
\hline
 Corridor & 4 & {\color{blue}18.78} & {\color{red}3} & {\color{blue}23.65}  & {\color{red}6} \\
\hline
\end{tabular}
\end{table}

\begin{table}[ht]
\caption{
Corridor - Our Approach
}
\begin{minipage}[b]{0.45\linewidth}
\centering
 \tiny
 \begin{tabular}{ |>{\centering\arraybackslash}m{0.25in} | >{\centering\arraybackslash}m{0.25in} | >{\centering\arraybackslash}m{0.25in} |>{\centering\arraybackslash}m{0.25in} |>{\centering\arraybackslash}m{0.25in} |}
\hline
{\bf { \it  PS}}  & PD I & PD II & PD III \\
\hline
SP I & 4.20 & 0 & 57.52 \\
\hline
SP II & 89.92 & 3.73 & 3.27 \\
\hline
SP III & 2.83 & 96.26 & 4.32 \\
\hline
SP IV & 3.03 & 0 & 34.87 \\
\hline
\end{tabular}
  
\label{fig:figure1}
\end{minipage}
\hspace{0.5cm}
\begin{minipage}[b]{0.45\linewidth}
\centering
 \tiny
 \begin{tabular}{ |>{\centering\arraybackslash}m{0.25in} | >{\centering\arraybackslash}m{0.25in} | >{\centering\arraybackslash}m{0.25in} |>{\centering\arraybackslash}m{0.25in} |>{\centering\arraybackslash}m{0.25in} |}
\hline
 {\bf { \it  AD}}	 & PD I & PD II & PD III \\
\hline
SP I & 10.02 & 0 & 89.97 \\
\hline
SP II & 95.82 & 1.87 & 2.29 \\
\hline
SP III & 5.55 & 88.88 & 5.55 \\
\hline
SP IV & 11.69 & 0 & 88.30 \\
\hline

\end{tabular}
  
\label{fig:figure2}
\end{minipage}
  
\end{table}

\begin{table}
\caption{
Corridor ORK 
}

\begin{minipage}[b]{.5\textwidth}
\centering
  \tiny
\begin{tabular}{ |>{\centering\arraybackslash}m{0.25in} |>{\centering\arraybackslash}m{0.25in} |>{\centering\arraybackslash}m{0.25in} | >{\centering\arraybackslash}m{0.25in} | >{\centering\arraybackslash}m{0.25in} |>{\centering\arraybackslash}m{0.25in} |>{\centering\arraybackslash}m{0.25in} |>{\centering\arraybackslash}m{0.25in} |}
\hline
{\bf { \it  PS}}  & PD I & PD II & PD III & PD IV & PD V & PD VI & PD VII\\
\hline
SP I & 98.86 & 0.38 & 10.63 & 56.10 & 3.48 & 11.31 & 7.19\\
\hline
SP II & 0 & 0 & 88.75 & 28.77 & 1.93 & 33.94 & 85.61\\
\hline
SP III & 1.13 & 99.61 & 0.60 & 8.01 & 89.92 & 20.43 & 7.19\\
\hline
SP IV & 0 & 0 & 0 & 7.10 & 4.65 & 34.30 & 0\\
\hline
\end{tabular}
  
  \label{tab:forpol}
  \vspace{1em} 
\end{minipage}
\quad
\begin{minipage}[b]{.5\textwidth}
    \tiny
    \centering

 \begin{tabular}{ |>{\centering\arraybackslash}m{0.25in} |>{\centering\arraybackslash}m{0.25in} |>{\centering\arraybackslash}m{0.25in} | >{\centering\arraybackslash}m{0.25in} | >{\centering\arraybackslash}m{0.25in} |>{\centering\arraybackslash}m{0.25in} |>{\centering\arraybackslash}m{0.25in} |>{\centering\arraybackslash}m{0.25in} |}
\hline
{\bf { \it  AD}}  & PD I & PD II & PD III & PD IV & PD V & PD VI & PD VII\\
\hline
SP I & 16.63 & 0.19 & 6.69 & 58.89 & 1.72 & 11.85 & 4.01\\
\hline
SP II & 0 & 0 & 32.77 & 17.73 & 0.56 & 20.87 & 28.05\\
\hline
SP III & 0.14 & 38.78 & 0.29 & 6.53 & 34.47 & 16.64 & 3.12\\
\hline
SP IV & 0 & 0 & 0 & 16.31 & 5.02 & 78.66 & 0\\
\hline
\end{tabular}

  \label{tab:revpol}
\end{minipage}
\end{table}

\begin{table}
\caption{
Corner
}
\begin{minipage}[b]{.5\textwidth}
  \centering
  \tiny
\begin{tabular}{ |>{\centering\arraybackslash}m{0.25in} |>{\centering\arraybackslash}m{0.25in} |>{\centering\arraybackslash}m{0.25in} | >{\centering\arraybackslash}m{0.25in} | >{\centering\arraybackslash}m{0.25in} |>{\centering\arraybackslash}m{0.25in} |>{\centering\arraybackslash}m{0.25in} |>{\centering\arraybackslash}m{0.25in} |}
\hline
{\bf { \it  PS}}  & PD I & PD II & PD III & PD IV & PD V\\
\hline
SP I & 50.58 & 100.00 & 10.13 & 21.38 & 96.15 \\
\hline
SP II & 0 & 0 & 7.61 & 76.66 & 3.84 \\
\hline
SP III & 49.41 & 0 & 82.24 & 1.95 & 0\\
\hline

\end{tabular}
  
 \vspace{1em} 
\end{minipage}\quad

\begin{minipage}[b]{.5\textwidth}
    \tiny
    \centering
 \begin{tabular}{ |>{\centering\arraybackslash}m{0.25in} |>{\centering\arraybackslash}m{0.25in} |>{\centering\arraybackslash}m{0.25in} | >{\centering\arraybackslash}m{0.25in} | >{\centering\arraybackslash}m{0.25in} |>{\centering\arraybackslash}m{0.25in} |>{\centering\arraybackslash}m{0.25in} |>{\centering\arraybackslash}m{0.25in} |}
\hline
{\bf { \it  AD}}  & PD I & PD II & PD III & PD IV & PD V\\
\hline
SP I & 9.94 & 41.02 & 18.85 & 18.74 & 11.42 \\
\hline
SP II & 0 & 0 & 17.31 & 82.12 & 0.5587 \\
\hline
SP III & 5.90 & 0 & 93.05 & 1.04 & 0\\
\hline
\end{tabular}
  
  \label{tab:revpol}
\end{minipage}
\end{table}

\begin{table}
\caption{
Corner-ORK
}
\begin{minipage}[b]{.45\textwidth}
  \centering
  \tiny
\begin{tabular}{ |>{\centering\arraybackslash}m{0.2in} |>{\centering\arraybackslash}m{0.2in} |>{\centering\arraybackslash}m{0.2in} | >{\centering\arraybackslash}m{0.2in} | >{\centering\arraybackslash}m{0.2in} |>{\centering\arraybackslash}m{0.2in} |>{\centering\arraybackslash}m{0.2in} |>{\centering\arraybackslash}m{0.2in}|>{\centering\arraybackslash}m{0.2in} |>{\centering\arraybackslash}m{0.2in} |}
\hline
{\bf { \it  PS}}  & PD I & PD II & PD III & PD IV & PD V & PD VI & PD VII  & PD VIII\\
\hline
SP I & 33.74 & 98.92 & 0.86 & 0 & 0.61 & 5.05 & 91.87 & 66.81 \\
\hline
SP II & 0 & 0 & 87.82 & 100.00 & 96.71 & 77.04 & 2.50 & 19.91 \\
\hline
SP III & 66.25 & 1.07 & 11.30 & 0 & 2.66 & 17.89 & 5.62 & 13.27 \\
\hline
\end{tabular}
  
  \label{tab:forpol}
  \vspace{1em} 
\end{minipage}\quad
\begin{minipage}[b]{.45\textwidth}
    \tiny
    \centering
 \begin{tabular}{ |>{\centering\arraybackslash}m{0.2in} |>{\centering\arraybackslash}m{0.2in} |>{\centering\arraybackslash}m{0.2in} | >{\centering\arraybackslash}m{0.2in} | >{\centering\arraybackslash}m{0.2in} |>{\centering\arraybackslash}m{0.2in} |>{\centering\arraybackslash}m{0.2in} |>{\centering\arraybackslash}m{0.2in}|>{\centering\arraybackslash}m{0.2in} |>{\centering\arraybackslash}m{0.2in} |}
\hline
{\bf { \it  AD}}  & PD I & PD II & PD III & PD IV & PD V & PD VI & PD VII  & PD VIII\\
\hline
SP I & 25.46 & 12.26 & 0.26 & 0 & 0.4 & 1.73 & 19.60 & 40.26 \\
\hline
SP II & 0 & 0 & 15.94 & 23.83 & 37.17 & 15.62 & 0.31 & 7.10 \\
\hline
SP III & 70.75 & 0.18 & 4.90 & 0 & 2.45 & 8.67 & 1.69 & 11.32 \\
\hline
\end{tabular}
  
  \label{tab:revpol}
\end{minipage}
\end{table}





%
\bibliographystyle{abbrv}
\bibliography{ref}

\end{document}